\documentclass{wscpaperproc}
\usepackage{latexsym}
\usepackage{graphicx}
\usepackage{mathptmx}
\usepackage[T1]{fontenc}

\usepackage{algorithm}
\usepackage{algpseudocode}
\usepackage{algpseudocode}
\usepackage{graphicx}
\usepackage{subfigure}
\usepackage{float}
\usepackage{multirow}
\usepackage{booktabs} % For better table lines
\usepackage{multicol} % For multicolumn cells
\usepackage{makecell}
\usepackage{amsmath}
\usepackage{amsfonts}
\usepackage{amssymb}
\usepackage{amsbsy}
\usepackage{amsthm}
\usepackage[pdftex,colorlinks=true,urlcolor=blue,citecolor=black,anchorcolor=black,linkcolor=black]{hyperref}

\newtheoremstyle{wsc}% hnamei
{3pt}% hSpace abovei
{3pt}% hSpace belowi
{}% hBody fonti
{}% hIndent amounti1
{\bf}% hTheorem head fontbf
{}% hPunctuation after theorem headi
{.5em}% hSpace after theorem headi2
{}% hTheorem head spec (can be left empty, meaning `normal')i

\theoremstyle{wsc}

\newtheorem{remark}{Remark}

\begin{document}

%***************************************************************************
% AUTHOR: AUTHOR NAMES GO HERE
% FORMAT AUTHORS NAMES Like: Author1, Author2 and Author3 (last names)
%
%		You need to change the author listing below!
%               Please list ALL authors using last name only, separate by a comma except
%               for the last author, separate with "and"
%

% setting up general page style
\pagestyle{fancyplain}

% setting up page style of first page
\thispagestyle{plain}
\firstPageHead{}

% setting up running header (authors) of subsequent pages
\chead{\fancyplain{}{\itshape Zehao Li and Yijie Peng}}

% setting up seperation parameters
%\headsep=72pt
\rhead{}
\cfoot{}
\renewcommand{\headrulewidth}{0pt} % (renewcommand needed in fancyhdr to remove top decorative line)
%\headrulewidth=0pt  % ("setlength" needed in fancyheading to remove top decorative line)

%%%%%%%%%%%%%%%%%%%%%%%%%%%%%%%%%%%%%%%%%%%%%%%%%%%%%%%%%%%%%%%%%%%%%%%%%%%%%%
%                                                                            %
%     THESE COMMANDS ARE REQUIRED TO WORK WITH WSC.BST TO MAKE BIBLIO     %
%                                                                            %
%%%%%%%%%%%%%%%%%%%%%%%%%%%%%%%%%%%%%%%%%%%%%%%%%%%%%%%%%%%%%%%%%%%%%%%%%%%%%%
\makeatletter
\let\@internalcite\cite
\def\cite{\def\@citeseppen{-1000}%
    \def\@cite##1##2{(##1\if@tempswa , ##2\fi)}%
    \def\citeauthoryear##1##2##3{##1 ##3}\@internalcite}
\def\citeNP{\def\@citeseppen{-1000}%
    \def\@cite##1##2{##1\if@tempswa , ##2\fi}%
    \def\citeauthoryear##1##2##3{##1 ##3}\@internalcite}
\def\citeN{\def\@citeseppen{-1000}%
%  Pierre L'Ecuyer's fix for multiple cite bug
%  Added by Paul J Sanchez on 4 October 2001
%   \def\@cite##1##2{##1\if@tempswa , ##2)\else{)}\fi}%
%   \def\citeauthoryear##1##2##3{##1 (##3}\@citedata}
    \def\@cite##1##2{##1\if@tempswa, ##2)\else{}\fi}%
    \def\citeauthoryear##1##2##3{##1 (##3)}\@citedata}
\def\citeA{\def\@citeseppen{-1000}%
    \def\@cite##1##2{(##1\if@tempswa , ##2\fi)}%
    \def\citeauthoryear##1##2##3{##1}\@internalcite}
\def\citeANP{\def\@citeseppen{-1000}%
    \def\@cite##1##2{##1\if@tempswa , ##2\fi}%
    \def\citeauthoryear##1##2##3{##1}\@internalcite}
\def\shortcite{\def\@citeseppen{-1000}%
    \def\@cite##1##2{(##1\if@tempswa , ##2\fi)}%
    \def\citeauthoryear##1##2##3{##2 ##3}\@internalcite}
\def\shortciteNP{\def\@citeseppen{-1000}%
    \def\@cite##1##2{##1\if@tempswa , ##2\fi}%
    \def\citeauthoryear##1##2##3{##2 ##3}\@internalcite}
\def\shortciteN{\def\@citeseppen{-1000}%
%  Pierre L'Ecuyer's fix for multiple cite bug
%  Added by Paul J Sanchez on 2 September 2002
%  should have caught this last year...
%   \def\@cite##1##2{##1\if@tempswa , ##2)\else{)}\fi}%
%   \def\citeauthoryear##1##2##3{##2 (##3}\@citedata}
% Shane G. Henderson fix for extra right bracket at end of optional material June 8, 2005
%    \def\@cite##1##2{##1\if@tempswa, ##2)\else{}\fi}%
    \def\@cite##1##2{##1\if@tempswa, ##2\else{}\fi}%
    \def\citeauthoryear##1##2##3{##2 (##3)}\@citedata}
\def\shortciteA{\def\@citeseppen{-1000}%
    \def\@cite##1##2{(##1\if@tempswa , ##2\fi)}%
    \def\citeauthoryear##1##2##3{##2}\@internalcite}
\def\shortciteANP{\def\@citeseppen{-1000}%
    \def\@cite##1##2{##1\if@tempswa , ##2\fi}%
    \def\citeauthoryear##1##2##3{##2}\@internalcite}
\def\citeyear{\def\@citeseppen{-1000}%
    \def\@cite##1##2{(##1\if@tempswa , ##2\fi)}%
    \def\citeauthoryear##1##2##3{##3}\@citedata}
\def\citeyearNP{\def\@citeseppen{-1000}%
    \def\@cite##1##2{##1\if@tempswa , ##2\fi}%
    \def\citeauthoryear##1##2##3{##3}\@citedata}
%
% \@citedata and \@citedatax:
%
% Place commas in-between citations in the same \citeyear, \citeyearNP,
% \citeN, or \shortciteN command.
% Use something like \citeN{ref1,ref2,ref3} and \citeN{ref4} for a list.
%
\def\@citedata{%
    \@ifnextchar [{\@tempswatrue\@citedatax}%
                  {\@tempswafalse\@citedatax[]}%
}

\def\@citedatax[#1]#2{%
\if@filesw\immediate\write\@auxout{\string\citation{#2}}\fi%
  \def\@citea{}\@cite{\@for\@citeb:=#2\do%
    {\@citea\def\@citea{, }\@ifundefined% by Young
       {b@\@citeb}{{\bf ?}%
       \@warning{Citation `\@citeb' on page \thepage \space undefined}}%
{\csname b@\@citeb\endcsname}}}{#1}}%

% don't box citations, separate with ; and a space
% also, make the penalty between citations negative: a good place to break.
%
\def\@citex[#1]#2{%
\if@filesw\immediate\write\@auxout{\string\citation{#2}}\fi%
  \def\@citea{}\@cite{\@for\@citeb:=#2\do%
    {\@citea\def\@citea{; }\@ifundefined% by Young
       {b@\@citeb}{{\bf ?}%
       \@warning{Citation `\@citeb' on page \thepage \space undefined}}%
{\csname b@\@citeb\endcsname}}}{#1}}%

% (from apalike.sty)
% No labels in the bibliography.
%
\def\@biblabel#1{}
\makeatother

%\newlength{\bibhang}
%\setlength{\bibhang}{2em}

% Indent second and subsequent lines of bibliographic entries. Taken
% from openbib.sty: \newblock is set to {}.
% \renewcommand{\refname}{REFERENCES}

\newdimen\bibindent
\bibindent=0.0em
% SEC: was \def\thebibliography#1{\section*{\refname\@mkboth
% SEC: was   {\uppercase{\refname}}{\uppercase{\refname}}}\list
\def\thebibliography#1{\section*{\refname}\list
   {}{\settowidth\labelwidth{[#1]}
   \leftmargin\parindent
   \itemindent -\parindent
   \listparindent \itemindent
   \itemsep 0pt
   \parsep 0pt}
   \def\newblock{}
   \sloppy
   \sfcode`\.=1000\relax}

           % Set up BiBTeX macros

% needed to make the tex document look more like the word counterpart :-(
\setlength{\baselineskip}{12.7pt}

% AUTHOR: Enter the title, all letters in upper case
\title{A New Stochastic Approximation Method for Gradient-based Simulated Parameter Estimation}

% AUTHOR: Enter the authors of the article, see end of the example document for further examples
\author{\begin{center}Zehao Li\textsuperscript{1,2} and Yijie Peng\textsuperscript{1,2}\\
[11pt]
\textsuperscript{1}Wuhan Institute for Artificial Intelligence, Guanghua School of Management, Peking University, Beijing, China,\\
\textsuperscript{2}Xiangjiang Laboratory, Changsha, Hunan, China,\end{center}
}

\maketitle

\vspace{-12pt}

\section*{ABSTRACT}
%This article addresses the challenge of parameter calibration in stochastic models where the likelihood function is not analytically available. We propose a gradient-based simulated parameter estimation framework, leveraging a multi-time scale algorithm that tackles the issue of ratio bias in both maximum likelihood estimation and posterior density estimation problems.  Numerical experiments show that our algorithm can improve the estimation accuracy and save computational costs.

This paper tackles the challenge of parameter calibration in stochastic models, particularly in scenarios where the likelihood function is unavailable in an analytical form. We introduce a gradient-based simulated parameter estimation framework, which employs a multi-time scale stochastic approximation algorithm. This approach effectively addresses the ratio bias that arises in both maximum likelihood estimation and posterior density estimation problems. The proposed algorithm enhances estimation accuracy and significantly reduces computational costs, as demonstrated through extensive numerical experiments. Our work extends the GSPE framework to handle complex models such as hidden Markov models and variational inference-based problems, offering a robust solution for parameter estimation in challenging stochastic environments.

\section{INTRODUCTION}
\label{sec:intro}
Parameter estimation is a vital aspect in fields like financial risk assessment and medical diagnosis, where it entails calibrating model parameters based on observed data. The frequentist perspective treats parameters as unknown constants, whereas the Bayesian perspective infers their posterior distribution. Important inference methods include maximum likelihood estimation (MLE), which provides consistency and asymptotic efficiency \shortcite{shao2003mathematical}, and posterior density estimation (PDE), which combines observed data with prior knowledge to ensure precise inference. Both techniques are extensively used in statistics and machine learning.

To solve the MLE, one relies on the analytical form of the logarithmic likelihood function. By substituting the observed data and solving for its maximum, the MLE can be derived. For PDE, the classical method involves variational inference \shortcite{blei2017variational}, which similarly requires an analytical form of the logarithmic likelihood. This method assumes a family of posterior distributions and minimizes the Kullback-Leibler divergence (KL divergence) to derive optimal posterior parameters. This paper focuses on stochastic models or simulators, which are characterized by system dynamics rather than explicit likelihood functions. Examples include Lindley’s recursion in queuing models, where the likelihood function of the output data does not have an analytical form, posing significant challenges for parameter calibration.

The MLE problem was first introduced and addressed by the gradient-based simulated maximum likelihood estimation (GSMLE) method in \shortciteN{Peng2020}. The Robbins-Monro algorithm, a classic stochastic approximation (SA) method \shortcite{Kushner1997StochasticAA}, is used to optimize unknown parameters for MLE. Specifically, let $Y$ denote the observed data, $\theta \in \mathbb{R}^d$ represent the parameter of interest, and $p$ stand for the unknown density. The gradient of the logarithmic likelihood function $\sum_{t=1}^{T} \log p(Y_t; \theta)$ with respect to $\theta$ can be expressed as a ratio:
\begin{equation}\label{eq1}
   \nabla_{\theta} \sum_{t=1}^{T} \log p(Y_t; \theta) = \sum_{t=1}^{T} \frac{\nabla_{\theta} p(Y_t; \theta)}{p(Y_t; \theta)}.
\end{equation} 
When no analytical form for the likelihood function is available, the generalized likelihood ratio (GLR) method is employed to obtain unbiased estimators for the density and its gradients \shortcite{peng2018new}. The GLR estimator provides unbiased estimators for "distribution sensitivities" as shown in \shortciteN{Lei2018ApplicationsOG}, achieving a square-root convergence rate \shortcite{glynn2021computing}. 

However, the gradient estimator of the logarithmic likelihood function presented in \shortciteN{Peng2020} is biased. Although the GLR estimator is unbiased, meaning we can obtain unbiased estimators $G_1(Y_t, \theta)$ and $G_2(Y_t, \theta)$ for $\nabla_{\theta} p(Y_t; \theta)$ and $p(Y_t; \theta)$ through the GLR method and Monte Carlo simulation, the ratio of these two unbiased estimators may not be unbiased. Therefore, when this ratio estimator is used in the Robbins-Monro algorithm, the update rule becomes:
\begin{equation}\label{single}
    \theta_{k+1} = \theta_k + \beta_k \sum_{t=1}^T \frac{G_1(Y_t, \theta_k)}{G_2(Y_t, \theta_k)},
\end{equation}
where the gradient term is biased, introducing a certain bias into the iterative results ($\beta_k$ is the step-size, satisfying specific step-size conditions). Additionally, the estimator in the denominator may cause numerical instability, resulting in inaccuracies in the MLE.

In the context of PDE, the computation of the log-likelihood function is similarly crucial. When the likelihood function does not have an analytical form, an estimator must be devised. In this simulation-based inference, also referred to as likelihood-free inference, various methods utilize neural networks to estimate likelihoods or posteriors that are otherwise infeasible to calculate \shortcite{glockler2022variational,pmlr-v97-greenberg19a,papamakarios2019sequential,NIPS2017_6f1d0705}. However, the likelihood functions inferred by neural networks tend to be biased. The integration of neural networks and the associated bias make these algorithms theoretically challenging. To simplify this and enable theoretical analysis, we frame this problem within the SA framework, utilizing unbiased GLR gradient estimators for the likelihood function as in the MLE case. Since the gradient estimator of the posterior density also involves Equation (\ref{eq1}), reducing the ratio bias in these stochastic models remains an open problem.

To tackle the ratio bias arising in both MLE and PDE, we propose a gradient-based simulated parameter estimation (GSPE) algorithm based on a multi-time scale (MTS) SA method \shortcite{Kushner1997StochasticAA,borkar2009stochastic}. The core idea is to treat both the parameters and the gradient of the logarithmic likelihood function together as parts of a stochastic root-finding problem. The method then approximates the solution by using two coupled iterations, with one component updated at a faster rate than the other. Specifically, we develop a recursive estimator that replaces the ratio form of the gradient estimator. This approach allows for incremental updates to the gradient estimators by averaging all available simulation data, thereby eliminating ratio bias throughout the iterative process. Similar approaches have been applied to quantile optimization, black-box CoVaR estimation, and dynamic pricing and replenishment problems \shortcite{hu2022stochastic,hu2024quantile,jiang2022quantile,Cao2023BlackboxCA,zheng2024dual}. 

Our approach, however, involves a more complex structure. The PDE problem is formulated as a nested simulation optimization problem within the variational inference framework. Minimizing KL divergence is equivalent to maximizing the evidence lower bound (ELBO), which is expressed as an expectation with respect to the unknown variational distribution. Consequently, the optimization objective is an expectation, and the sample average approximation (SAA) method is applied to obtain an unbiased gradient estimator for the ELBO, forming the outer layer of the simulation. Meanwhile, the intractable likelihood within this expectation is estimated using the inner-layer simulation and unbiased GLR estimators. A nested MTS algorithm is designed to address ratio bias, solving the nested simulation optimization problem.

Furthermore, the MLE in hidden Markov models (HMM) is also a difficult problem. The likelihood function in HMMs is a high-dimensional integral over the hidden states, which does not have a closed form. Estimating the gradient of this likelihood becomes problematic. Sequential Monte Carlo (SMC), also known as particle filtering, is a standard method for handling HMMs \shortcite{wills2023sequential,doucet2001sequential}. We find that the proposed algorithm can be applied to this complex scenario in conjunction with SMC.

In this paper, we introduce a new GSPE framework that can asymptotically eliminate ratio bias in parameter estimation without requiring an analytical likelihood function. The MTS algorithm is employed in the MLE problem to enhance estimation accuracy and reduce computational cost. The GSPE framework also incorporates a nested MTS algorithm to solve the PDE problem in conjunction with variational inference. MLE for HMMs is also addressed as a specific case. This work extends the previous GSPE framework in \shortciteN{li2024eliminating} to HMMs, with all theoretical results omitted.

The paper is organized as follows: Section \ref{section2} provides the necessary background and introduces the GSPE algorithm framework for both MLE and PDE. Section \ref{sec5} presents numerical results, and Section \ref{sec6} concludes the paper.

\section{Problem Setting and Algorithm Design}\label{section2}
This section outlines the fundamental problem settings within the GSPE framework. To address the issue of ratio bias in the MLE problem, we propose the MTS algorithm in Section \ref{section2.1}. Additionally, a nested MTS approach is introduced to tackle the PDE problem in Section \ref{section2.2}. Furthermore, MLE for HMMs is discussed in Section \ref{sec2.3}.
\subsection{Maximum Likelihood Estimation}\label{section2.1}
    
    Considering a stochastic model, let $X$ be a random variable with density function $f(x,\theta)$ where $\theta \in \mathbb{R}^d$ is the parameter with feasible domain $\Theta\subset\mathbb{R}^d$. Another random variable $Y$ is defined by the relationship
    %\begin{equation*}%\label{1}
        $Y = g(X,\theta)$,
    %\end{equation*}
    where $g$ is known in analytical form. In this model, $Y$ is observable with $X$ being latent. Our objective is to estimate the parameter $\theta$ based on the observed data $y:=\{Y_t\}_{t=1}^T$. 
    
    In a special case where $X$ is one-dimensional with density $f(x)$, and $g$ is invertible with a differentiable inverse with respect to the $y$, a standard result in probability theory allows the density of $Y_t$ to be expressed in closed form as: $p(y;\theta) = f(g^{-1}(y;\theta))|\frac{d}{dy}g^{-1}(y;\theta)|.$
    %$$p(y;\theta) = f(g^{-1}(y;\theta))\bigg|\frac{d}{dy}g^{-1}(y;\theta)\bigg|.$$
    However, the theory developed in this paper does not require such restrictive assumptions. Instead, we only assume that $g$ is differentiable with respect to $x$ and that its gradient is non-zero a.e. 
    
    Under this weaker condition, even though the analytical forms of $f$ and $g$ are known, the density of $Y$ may still be unknown. In this case, the likelihood function for $Y$ can only be expressed as:
    \begin{equation}\label{2}
    L_T(\theta) := \sum\limits_{t=1}\limits^{T}\log p(Y_t;\theta).
\end{equation}
To maximize $L_{T}(\theta)$, we compute the gradient of the log-likelihood: 
\begin{equation}\label{equation3}
    \nabla_{\theta}L_T(\theta) = \sum\limits_{t=1}\limits^{T}\frac{\nabla_{\theta}p(Y_t;\theta)}{p(Y_t;\theta)}.
\end{equation}

    Suppose we have unbiased estimators for $\nabla_{\theta}p(Y_t;\theta)$ and ${p(Y_t;\theta)}$ for every $\theta$ and $Y_t$. While these individual estimators are unbiased, the ratio of two unbiased estimators may introduce bias. To distinguish between approaches, we refer to the previous algorithm using the plug-in estimator from Equation (\ref{single}) as the single time scale (STS) algorithm \shortcite{Peng2020}. To address this issue, we adopt an MTS framework that incorporates the gradient estimator into the iterative process, aiming for more accurate optimization results. Specifically, let $G_1(X,y,\theta)$ and $G_2(X,y,\theta)$ represent unbiased estimators obtained via Monte Carlo simulation:
\begin{equation}\label{G1G2}
    G_1(X,Y_t,\theta) = \frac{1}{N}\sum_{i=1}^NG_1(X_i,Y_t,\theta),\quad G_2(X,Y_t,\theta) = \frac{1}{N}\sum_{i=1}^NG_2(X_i,Y_t,\theta),
\end{equation}
    such that $$\mathbb{E}_X[G_1(X,Y_t,\theta)] = \nabla_{\theta}p(Y_t;\theta), \quad \mathbb{E}_X[G_2(X,Y_t,\theta)] = p(Y_t;\theta).$$ 

The forms of $G_1$ and $G_2$ can be derived by GLR estimators \shortcite{Peng2020}.  Alternative single-run unbiased estimators for $G_1$ and $G_2$ can also be obtained via the conditional Monte Carlo method, as described in \shortcite{fu2009conditional}. We propose the iteration formulae for the MTS algorithm as follows:
    \begin{equation}\label{eq3}
        D_{k+1} = D_k + \alpha_k(G_{1,k}(X,Y,\theta_k)-G_{2,k}(X,Y,\theta_k)D_k),
    \end{equation}
    \begin{equation}\label{eq2}
        \theta_{k+1} = \Pi_{\Theta}(\theta_k + \beta_k E D_{k}),
    \end{equation}
     where $\Pi_{\Theta}$ is the projection operator that maps each iteratively obtained $\theta_k$ onto the feasible domain $\Theta$. The algebraic notations are as follows. $G_{1,k}(X,Y,\theta_k)$ represents the combination of all estimators  $G_{1}(X,Y_t,\theta_k)$ under every observation $Y_t$, forming a column vector with $T\times d$ dimensions. $G_{2,k}(X,Y,\theta_k)$ is also the combination of all estimators $G_{2}(X,Y_t,\theta_k)$ under every observation $Y_t$. That is to say, $G_{2,k}(X,Y,\theta_k) = diag \{G_{2}(X,Y_1,\theta_k)I_d,\cdots,G_{2}(X,Y_T,\theta_k)I_d\} = diag \{G_{2}(X,Y_1,\theta_k),\cdots,G_{2}(X,Y_T,\theta_k)\}\otimes I_d$, which is a diagonal matrix with $T\times d$ rows and $T\times d$ columns. $\otimes$ stands for Kronecker product and $I_d$ denotes the $d$-dimensional identity matrix. The constant matrix $E=[I_d,I_d,\cdots,I_d] = e^T\otimes I_d$ is a block diagonal matrix with $d$ rows and $T\times d$ column, where $e$ is a  column vector of ones. This matrix reshapes the long vector $D_k$ to match the structure of Equation (\ref{equation3}), the summation of $T$ $d$-dimensional vectors. 
     
     In these two coupled iterations, $\theta_k$ is the parameter being optimized in the MLE process, as in Equation (\ref{single}). The additional iteration for $D_k$  tracks the gradient of the log-likelihood function, mitigating ratio bias and numerical instability caused by denominator estimators. These two iterations operate on different time scales, with distinct update rates. Ideally, one would fix $\theta$, run iteration (\ref{eq3}) until it converges to the true gradient, and then use this limit in iteration (\ref{eq2}). However, such an approach is computationally inefficient. Instead, these coupled iterations are executed interactively, with iteration (\ref{eq3}) running at a faster rate than (\ref{eq2}), effectively treating $\theta$ as fixed in the second iteration. This time-scale separation is achieved by ensuring that the step sizes satisfy: 
     $\frac{\beta_k}{\alpha_k} \rightarrow 0$ as $k$ tends to infinity. This design allows the gradient estimator's bias to average out over the iteration process, enabling accurate results even with a small Monte Carlo sample size $N$ in Equation (\ref{G1G2}).  Ultimately, $E D_{k}$ converges to zero, and $\theta$ converges to its optimal value.  The MTS framework for MLE is summarized as follows. 
\begin{algorithm}[h!]
\small
   \caption{(MTS for MLE)}
   \label{algor:1}
   \begin{algorithmic}[1]
   \State Input: data$\{Y_t\}_{t=1}^T$, initial iterative values $\theta_0$, $D_0$, number of samples $N$, iterative steps $K$, the step-sizes $\alpha_k$, $\beta_k$.
   \For {$k \text{ in } 0: K-1$}   
   \State For $i=1:N$, sample $X_i$ and get unbiased estimators $G_{1,k}(X_i,Y,\theta_k)$, $G_{2,k}(X_i,Y,\theta_k)$.
   \State Do the iterations: 
           $D_{k+1} = D_k + \alpha_k(G_{1,k}(X,Y,\theta_k)-G_{2,k}(X,Y,\theta_k)D_k), \quad
   \theta_{k+1} = \Pi_{\Theta}(\theta_k +\beta_k E D_{k}).$
   \EndFor
   \State Output: $\theta_{K}$.
   \end{algorithmic}
\end{algorithm}
\subsection{Posterior Density Estimation}\label{section2.2}

We now  turn to the problem of estimating the posterior distribution of the parameter $\theta$ in the stochastic model $Y=g
(X,\theta)$, where the analytical likelihood is unknown. The posterior distribution is defined as
\begin{equation*}
    p(\theta|y) = \frac{p(\theta)p(y|\theta)}{\int p(\theta)p(y|\theta)d\theta},
\end{equation*}
where $p(\theta)$ is the known prior distribution, and $p(y|\theta)$ is the conditional density function that lacks an analytical form but can be estimated using an unbiased estimator. The denominator is a challenging normalization constant to handle and variational inference is a practical approach.

In the variational inference framework, we approximate the posterior distribution $p(\theta|y)$ using a tractable density $q_{\lambda}(\theta)$ with a variational parameter $\lambda$ to approximate. The collection $\{q_{\lambda}(\theta)\}$ is called the variational distribution family, and our goal is to find the optimal $\lambda$ by minimizing the KL divergence between tractable variational distribution $q_{\lambda}(\theta)$ and the true posterior $p(\theta|y)$:
$$KL(\lambda) = KL(q_{\lambda}(\theta)\Vert p(\theta|y))=\mathbb{E}_{q_{\lambda}(\theta)}[\log q_{\lambda}(\theta)-\log p(\theta|y)].$$
It is well known that minimizing KL divergence is equivalent to maximizing the ELBO, an expectation with respect to variational distribution $q_{\lambda}(\theta)$: $$L(\lambda) = \log p(y) - KL(\lambda) = \mathbb{E}_{q_{\lambda}(\theta)}[\log p(y|\theta) + \log p(\theta) - \log q_{\lambda}(\theta)].$$ The problem is then reformulated as   $$ \lambda^{*} = \arg\max\limits_{\lambda \in \Lambda}{L(\lambda)},$$ where $\Lambda$ is the feasible region of $\lambda$. It is essential to estimate the gradient of ELBO, which is an important problem in the field of machine learning and also falls under the umbrella of simulation optimization.  Common methods for deriving gradient estimators include the score function method \shortcite{ranganath2014black} and the re-parameterization trick \shortcite{kingma2013auto,rezende2014stochastic}. In the simulation literature, these methods are also referred to as the likelihood ratio (LR) method and infinitesimal perturbation analysis (IPA) method, respectively \shortcite{FU2006575}.

In this paper, $p(y|\theta)$ is estimated by simulation rather than computed precisely, inducing bias to the $\log p(y|\theta)$ term in LR method. Furthermore, the LR method is prone to high variance \shortcite{rezende2014stochastic}, making the re-parameterization trick a preferred choice.

Assume a variable substitution involving 
$\lambda$, such that $\theta = \theta(u;\lambda) \sim q_{\lambda}(\theta)$, where $u$ is a random variable independent of $\lambda$ with density $p_0(u)$. This represents a re-parameterization of $\theta$, where the stochastic component is incorporated into $u$, while the parameter $\lambda$ is isolated. Allowing the interchange of differentiation and expectation \shortcite{glasserman1990gradient}, we obtain
\begin{equation}\label{equation2}
    \begin{aligned}
        \nabla_{\lambda}L(\lambda) =& \nabla_{\lambda}\mathbb{E}_{q_{\lambda}(\theta)}[\log p(y|\theta) + \log p(\theta) - \log q_{\lambda}(\theta)] \\ =& \nabla_{\lambda}\mathbb{E}_{u}[\log p(y|\theta(u;\lambda)) + \log p(\theta(u;\lambda)) - \log q_{\lambda}(\theta(u;\lambda))] \\ =& \mathbb{E}_{u}[\nabla_{\lambda}\theta(u;\lambda)\cdot(\nabla_{\theta}\log p(y|\theta) + \nabla_{\theta}\log p(\theta) - \nabla_{\theta}\log q_{\lambda}(\theta))].
    \end{aligned}
\end{equation}

In Equation (\ref{equation2}), the Jacobi term $\nabla_{\lambda}\theta(u;\lambda)$, prior term $\log p(\theta)$ and variational distribution term $\log q_{\lambda}(\theta)$ are known. Therefore, the focus is on the term involving the intractable likelihood function. Similar to the MLE case, the term $\nabla_{\theta}\log p(y|\theta) = \frac{\nabla_{\theta}p(y|\theta)}{p(y|\theta)}$ contains the ratio of two estimators, which introduces bias. 

The problem differs in two aspects. First, the algorithm no longer iterates over the parameter $\theta$ to be estimated but over the variational parameter $\lambda$, which defines the posterior distribution. This shifts the focus from point estimation to function approximation, aiming to identify the best approximation of the true posterior from the variational family ${q_{\lambda}(\theta)}$. Second, this becomes a nested simulation problem because the objective is ELBO, an expectation over a random variable $u$. Estimating its gradient requires an additional outer-layer simulation using SAA. In the outer layer simulation, we sample $u$ to get the different $\theta$, representing various scenarios. For each $\theta$, the likelihood function and its gradient are estimated using the GLR method as in the MLE case, incorporating the MTS framework to reduce ratio bias. After calculating the part inside the expectation in Equation (\ref{equation2}) for every sample $u$, we average the results with respect to $u$ to get the estimator of the gradient of ELBO. 
 
 Note that the inner layer simulation for term $\nabla_{\theta}\log p(y|\theta) = \frac{\nabla_{\theta}p(y|\theta)}{p(y|\theta)}$ depends on $u$, 
 so we need to fix outer layer samples $\{u_m\}_{m=1}^M$ at the beginning of the algorithm. Similar to the MLE case, $M$ parallel gradient iteration processes are defined as blocks $\{D_{k,m}\}_{m=1}^M$, where  $D_{k,m}$ tracks the gradient of the likelihood function $\nabla_{\theta}\log p(y|\theta(u_m;\lambda_k))$ for every outer layer sample $u_m$. The optimization process of $\lambda$ depends on the gradient of ELBO in Equation (\ref{equation2}), which is estimated by averaging over these $M$ blocks. An additional error arises between the true gradient of ELBO and its estimator due to outer-layer simulation. Unlike Algorithm \ref{algor:1}, this approach involves a nested simulation optimization structure, where simulation and optimization are conducted simultaneously. 
 
 The nested MTS algorithm framework for the PDE problem is shown as Algorithm \ref{algor:2}. $G_{1,k}(X,Y,\theta_{k,m})$ and $G_{2,k}(X,Y,\theta_{k,m})$ could be unbiased GLR estimators. The matrix dimensions are consistent with those in the MLE case. The iteration for $D_{k,m}$ resembles the MLE case, except for the parallel blocks.  The iteration for $\lambda_k$ corresponds to the gradient  $\nabla_{\lambda}L(\lambda)$ in Equation (\ref{equation2}). 

\begin{algorithm}[h] 
\small
   \caption{(Nested MTS for PDE)}
   \label{algor:2}
   \begin{algorithmic}[1]
   \State Input: data $\{Y_t\}_{t=1}^T$, prior $p(\theta)$, iteration initial value $\lambda_0$ and $D_0$, iteration times $K$, number of outer layer samples $M$, number of inner layer samples $N$, step-sizes $ \alpha_k$, $\beta_k$.
   \State Sample $\{u_m\}_{m=1}^M$ from $p_0(u)$ as outer layer samples.
   \For {$k \text{ in } 0: K-1$}
   \State $\theta_{k,m} = \theta(u_m;\lambda_k)$, for $m=1:M$;
   \State Sample $\{X_i\}_{i=1}^N$ and get the inner unbiased layer estimators $G_{1,k}(X,Y,\theta_{k,m})$, $G_{2,k}(X,Y,\theta_{k,m})$, for $i=1:N$ and  $m=1:M$;
   \State Do the iterations: $$D_{k+1,m} = D_{k,m} + \alpha_k(G_{1,k}(X,Y,\theta_{k,m})-G_{2,k}(X,Y,\theta_{k,m})D_{k,m}).$$
   $$\lambda_{k+1} = \Pi_{\Lambda}\bigg(\lambda_k + \beta_k \frac{1}{M}\sum_{m=1}^M\bigg(\nabla_{\lambda}\theta(u;\lambda)\bigg|_{(u;\lambda)=(u_m;\lambda_k)}\bigg(ED_{k,m} + \nabla_{\theta}\log p(\theta_{k,m}) - \nabla_{\theta}\log q_{\lambda}(\theta_{k,m})\bigg)\bigg)\bigg).$$
  %\State The ELBO is $$\hat{L}(\lambda_k) = \frac{1}{M}\sum_{m=1}^M\bigg(\log p(Y|\theta_{k,m}) + \log p(\theta_{k,m}) - \log q_{\lambda}(\theta_{k,m})\bigg).$$     
   \EndFor
   \State Output: $\lambda_{K}$.
   \end{algorithmic}
\end{algorithm}

The following remark highlights the advantage of the MTS algorithm compared to the STS algorithm. 
\begin{remark}\label{remark2}
        In the PDE case, the corresponding iterative process of STS is as below:
    \begin{equation}\label{single2}
        \lambda_{k+1} = \Pi_{\lambda}\bigg(\lambda_k + \beta_k \frac{1}{M}\sum_{m=1}^M\bigg(\nabla_{\lambda}\theta(u_m;\lambda_k)\bigg(\sum_{t=1}^T\frac{G_{1}(X,Y_t,\theta_{k,m})}{G_{2}(X,Y_t,\theta_{k,m})} + \nabla_{\theta}\log p(\theta_{k,m}) - \nabla_{\theta}\log q_{\lambda}(\theta_{k,m})\bigg)\bigg)\bigg).
    \end{equation}
 In this previous way, we do not use $D_k$ to track the gradient but plug in the ratio of two estimators whose bias may not be negligible if $N$ is not large enough. Moreover, the estimator in the denominator makes the algorithm numerically unstable. Therefore, the gradient estimated in this algorithm is not precise so the optimization process is impacted. In Section \ref{sec5}, we will find that the STS algorithm does not perform as well as MTS.
    \end{remark}

\subsection{MLE for the Hidden Markov Models}\label{sec2.3}
Generally, an HMM can be specified by the following general state space model: for $t = 1,\cdots, T$,
\begin{equation*}
    Y_t = g(W_t;S_t,\theta), \quad S_t = h(V_t;S_{t-1},\theta),
\end{equation*}
where $\{V_t\}_{t=1}^T$ are i.i.d. random variables driving the hidden underlying Markov chain $\{S_t\}_{t=1}^T$ with initial state $S_0$. The model dynamics is governed by some parameter $\theta$ belonging to some parameter space $\Theta$. $\{W_t\}_{t=0}^T$ are i.i.d. random variables introducing interference to the unobservable state $S_t$ of the Markov chain. Only $\{Y_t\}_{t=1}^T$ are observable. For given observation data $\{Y_t\}_{t=1}^T$, the log-likelihood of observations following an HMM is given by
\begin{equation}
    L_T(\theta) \doteq \log \mathbb{E}\bigg[\prod_{t=1}^Tp_{\theta}(Y_t;S_t)\bigg],
\end{equation}
where $p_{\theta}(\cdot;S_t)$ is the conditional density of observation $Y_t$ on hidden state $S_t$ and the expectation is taken w.r.t $S_t$. The asymptotic properties of the MLE for an HMM are similar to the i.i.d. case and can be found
in \shortciteN[chap. 6]{Capp2010InferenceIH}.

To derive the gradient estimator $\nabla_{\theta}L_T(\theta)$ and reduce its variance, we need to construct a consecutive update of the prior distribution by incorporating information from observations sequentially and sample from the posterior. We decompose the log-likelihood into a sum of log conditional expectations: 
\begin{equation*}
    L_T(\theta) = \sum_{t=0}^{T-1}\log \bigg(\mathbb{E}\bigg[\prod_{l=1}^{t+1}p(Y_l;S_l,\theta)\bigg]\bigg/\mathbb{E}\bigg[\prod_{l=1}^{t}p(Y_l;S_l,\theta)\bigg]\bigg) \doteq  \sum_{t=0}^{T-1}\log \pi_{t+1|t}(p_{\theta}(S_{t+1};Y_{t+1})),
\end{equation*}
where $\prod_{l}^0\doteq 1$ and $\pi_{t+1|t}(p_{\theta}(S_{t+1};Y_{t+1})) = \mathbb{E}[p_{\theta}(S_{t+1};Y_{t+1})|Y_{1:t}]$. 
The gradient of the log-likelihood $L_{T}(\theta)$ becomes
\begin{equation}\label{gradient}
    \nabla_{\theta}L_{T}(\theta) = \sum_{t=0}^{T-1}\frac{\nabla \pi_{t+1|t}(p_{\theta}(S_{t+1};Y_{t+1}))}{\pi_{t+1|t}(p_{\theta}(S_{t+1};Y_{t+1}))}.
\end{equation}

Take the derivative and we can get
\begin{equation*}
    \nabla\pi_{t+1|t}(p_{\theta}(S_{t+1};Y_{t+1})) = \frac{\nabla\mathbb{E}[p_{\theta}(S_{t+1};Y_{t+1})\prod_{k=0}^tp_{\theta}(S_{k};Y_{k})]}{\mathbb{E}[\prod_{k=0}^tp_{\theta}(S_{k};Y_{k})]} - \pi_{t+1|t}(p_{\theta}(S_{t+1};Y_{t+1}))\frac{\nabla\mathbb{E}[\prod_{k=0}^tp_{\theta}(S_{k};Y_{k})]}{\mathbb{E}[\prod_{k=0}^tp_{\theta}(S_{k};Y_{k})]},
\end{equation*}
where
\begin{equation*}
\begin{aligned}
\nabla\mathbb{E}\bigg[\prod_{k=0}^{t+1}p_{\theta}(S_k;Y_k)\bigg] = \mathbb{E}\bigg[\bigg(\nabla p_{\theta}(S_{t+1};Y_{t+1}) +  p_{\theta}(S_{t+1};Y_{t+1})\sum_{k=0}^t\frac{\nabla p_{\theta}(S_{k};Y_{k})}{p_{\theta}(S_{k};Y_{k})}\bigg)\prod_{k=0}^Tp_{\theta}(S_{k};Y_{k})\bigg].
\end{aligned}
\end{equation*}
We define $Z_t = \frac{\partial S_t}{\partial \theta}$ and set the augmented Markov chain $(S_t,Z_t,W_t)_{t\ge 0}$ by the following recursive relationship:
$$S_{t+1} = h(V_{t+1};S_{t},\theta), \ Z_{t+1} = \frac{\partial S_{t+1}}{\partial \theta} = \frac{\partial h}{\partial \theta} + \frac{\partial h}{\partial S_t}Z_t,$$  
\begin{equation}\label{eq_w}
    W_{t+1} = W_t+\frac{\frac{\partial}{\partial S_t}p_{\theta}(S_{t+1};Y_{t+1})Z_{t+1} + \frac{\partial}{\partial \theta}p_{\theta}(S_{t+1};Y_{t+1})}{p_{\theta}(S_{t+1};Y_{t+1})}.
\end{equation}
Then  based on the SMC method, $\pi_{t+1|t}(p_{\theta}(S_{t+1};Y_{t+1}))$ can be estimated by the consistent estimator $\frac{1}{J}\sum_{j=1}^Jp_{\theta}(\hat{S}_{t+1}^j;Y_{t+1})$. And $\nabla \pi_{t+1|t}(p_{\theta}(S_{t+1};Y_{t+1}))$ can be estimated by the consistent estimator $\sum_{j=1}^J\nabla_{\theta} p_{\theta}(\hat{S}_{t}^j;Y_{t}) + p_{\theta}(\hat{S}_{t}^j;Y_{t})(W_{t-1}^j-\frac{1}{J}\sum_{j'}W_{t-1}^{j'})$. We deduce the IPA estimator of $\nabla L_T(\theta)$:
\begin{equation}\label{IPA}
\sum_{t=1}^T\frac{\sum_{j=1}^J\nabla_{\theta} p_{\theta}(\hat{S}_{t}^j;Y_{t}) + p_{\theta}(\hat{S}_{t}^j;Y_{t})(W_{t-1}^j-\frac{1}{J}\sum_{j'}W_{t-1}^{j'})}{\sum_{j=1}^Jp_{\theta}(\hat{S}_{t}^j;Y_{t})},
\end{equation}
where $(\hat{S}_{t}^j,Z_t^j,W_{t-1}^j)$ are particles derived by using a SMC algorithm on the augmented Markov chain. 

Noting that Equation (\ref{IPA}) contains the ratio of two estimators, we apply the GSPE algorithm and design an additional iteration $\{D_k\}$ to track this gradient in Equation (\ref{gradient}). The algorithm framework we propose is shown in Algorithm \ref{algo1}. In the simulation, we sample different hidden states $\hat{S}_t^j$ for $j = 1,\cdots, J$ by transition density $p(S_t^j|S_{t-1}^j,\theta)$ to obtain different particles for every observation $t = 1,\cdots, T$. Then we use the observation density $p(Y_t|S_t^j,\theta)$ to calculate the likelihood function and its derivatives in the corresponding hidden states, and assign different weights to different particles by comparing them with the real observations. Afterward, resampling is performed to prevent particle degradation caused by uneven weights. Then the numerator and the denominator of Equation (\ref{gradient}) can be approximated by the weighted average of all the $J$ particles through Equation (\ref{IPA}) separately, denoted as $G_{1,k}$ and $G_{2,k}$.

These two iterations operate on different time scales, with distinct update rates: 
         $\frac{\beta_k}{\alpha_k} \rightarrow 0$ as $k$ tends to infinity. The update rule for $\{D_k\}$ is based on a fixed-point principle, mitigating ratio bias and numerical instability caused by denominator estimators. This design allows the gradient estimator's bias to average out over the iteration process, enabling accurate results even with a small particle number $J$.  The estimation of the gradient of the likelihood function is finally obtained by the gradient ascent in the second time scale. The algorithm framework we propose is as follows. 

\begin{algorithm}[h!]
\footnotesize
   \caption{(MTS for the MLE in HMMs)}
   \label{algo1}
   \begin{algorithmic}[1]
   \State Input: data $\{Y_t\}_{t=1}^T$, initial iterative values $\theta_0$, $D_0$, number of particles $J$, iterative steps $K$, the stepsize $\alpha_k$, $\beta_k$.
   \State initialization: $D_0=0$, $w_0^j = 1/J$, for every $j =1,\cdots,J$.
   \For {$k \text{ in } 1: K$}   
   \For {$t \text{ in } 1: T$} 
   \State sample $V_t^j$ and get new state by $S_t^j = h(V_t^j;S_{t-1}^{j}, \theta_k)$, $j = 1,\cdots,J$;
   \State calculate the conditional density of every particle and their derivative: for $j =1,\cdots,J,$$$\Phi_{1,t,k}^j(Y_t,S_t^j,\theta_k)  = p(Y_t|S_t^j,\theta_k), \quad \Phi_{2,t,k}^j(Y_t,S_t^j,\theta_k) = \frac{\partial p(Y_t|s,\theta)}{\partial \theta}\bigg|_{s=S_t^j,\theta=\theta_k},$$
   $$\Phi_{3,t,k}^j(Y_t,S_t^j,\theta_k) = \frac{\partial p(Y_t|s,\theta)}{\partial s}\bigg|_{s=S_t^j,\theta=\theta_k},\quad \Phi_{4,t,k}^j = \frac{\partial h(V_t^j;s,\theta)}{\partial\theta}\bigg|_{s=S_{t-1}^j,\theta=\theta_k}.$$  
   \State calculate the estimator of the numerator and denominator of the SMC:
   \begin{equation*}
       \begin{aligned}      G_{1,k}(Y,\theta_k)(t)=\sum_{j=1}^J\Phi_{1,t,k}^j\Bigg(\sum_{l=1}^t\frac{\Phi_{2,l,k}^j + \Phi_{3,l,k}^j \Phi_{4,l,k}^j}{\Phi_{1,l,k}^j} - \frac{1}{J}\sum_{j'=1}^J\sum_{l=1}^{t-1}\frac{\Phi_{2,l,k}^{j'}
             + \Phi_{3,l,k}^{j'} \Phi_{4,l,k}^{j'}}{\Phi_{1,l,k}^{j'} }w_t^{j'}\Bigg) w_t^j, \quad
             G_{2,k}(Y,\theta_k)(t) = \sum_{j=1}^J\Phi_{1,t,k}^j w_t^j.
       \end{aligned}
   \end{equation*}
   \State If $ESS := \bigg(\sum_{j=1}^J(w_t^j)^2 > J/3 \bigg)^{-1} $, calculate the importance weight of every particle and update it:
   $$w_t^j = \frac{\Phi_{1,t,k}^j w_t^j}{\sum_{j=1}^J\Phi_{1,t,k}^j w_t^j}, \quad j =1,\cdots,J,$$
   \State else: using polynomial resampling to resample $S_t^j$ with probability of $w_t^j$, i.e. sample $S_t^j = S_t^{\xi_j}$ with $\xi_j = i \in {1,\cdots,J}$ w.p. $w_t^j$.  Then reset the $w_t^j = 1/J$.
   \EndFor
   \State Do the iterations: $$D_{k+1} = D_k + \alpha_k(G_{1,k}(Y,\theta_k) - G_{2,k}(Y,\theta_k) \cdot D_k),\quad \theta_{k+1} = \Pi_{\Theta}(\theta_k + \beta_k ED_{k}).$$
   \EndFor
   \State Output: $\theta_{K+1}$.
    \end{algorithmic}
\end{algorithm}

\section{Numerical Experiments}\label{sec5}
In this section, we demonstrate the application of the GSPE algorithm framework, comprising two specific algorithms, to various cases.  Section \ref{sec5.1} addresses the MLE case, while Section \ref{sec5.2} focuses on the PDE case. Section \ref{5.3} is a simple HMM case.

\subsection{MLE Case}\label{sec5.1}
We apply Algorithm \ref{algor:1} to evaluate the MTS framework in the MLE setting. Consider i.i.d. observations generated by the data-generating process $Y_t = g(X_t;\theta) = X_{1,t}+\theta X_{2,t},$ where $X_{1,t}, X_{2,t} \sim N(0,1)$ are independent. $Y_t$ is observable but $X_t$ is latent variable. The goal is to estimate $\theta$ based on observation $\{Y_t\}_{t=1}^T$. For this example, the MLE has an analytical form: $\hat{\theta} = \sqrt{\frac{1}{T}\sum\limits_{t=1}^T Y_t^2-1}.$

\begin{figure}[h]
  \centering
  \caption{Trajectories of MTS and STS with different sample sizes based on 100 independent experiments} 
  \subfigure[Convergence curves with $N = 10^4$]{
    \centering
    \includegraphics[width=7.5cm]{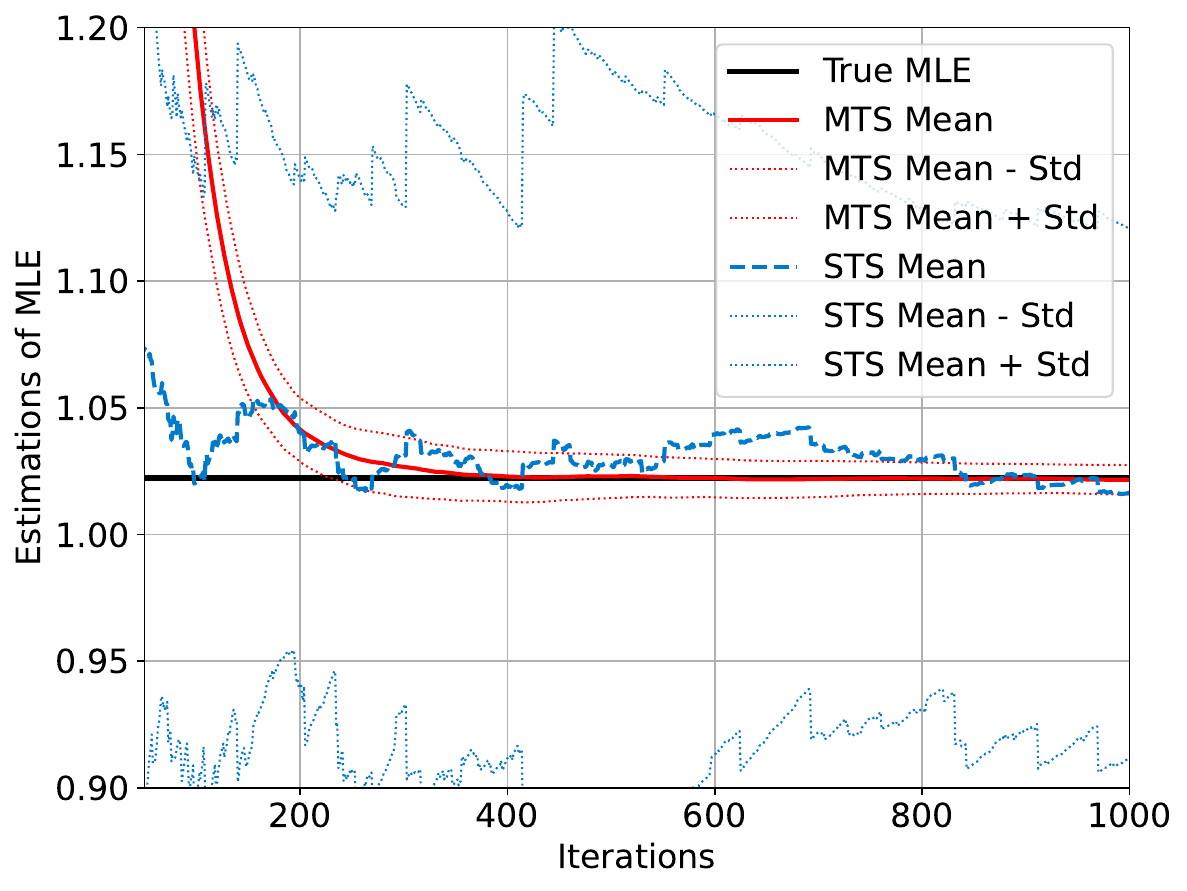}
    \label{ex1_2_1}
  }
  \subfigure[Convergence curves with $N = 10^5$]{
    \centering
    \includegraphics[width=7.5cm]{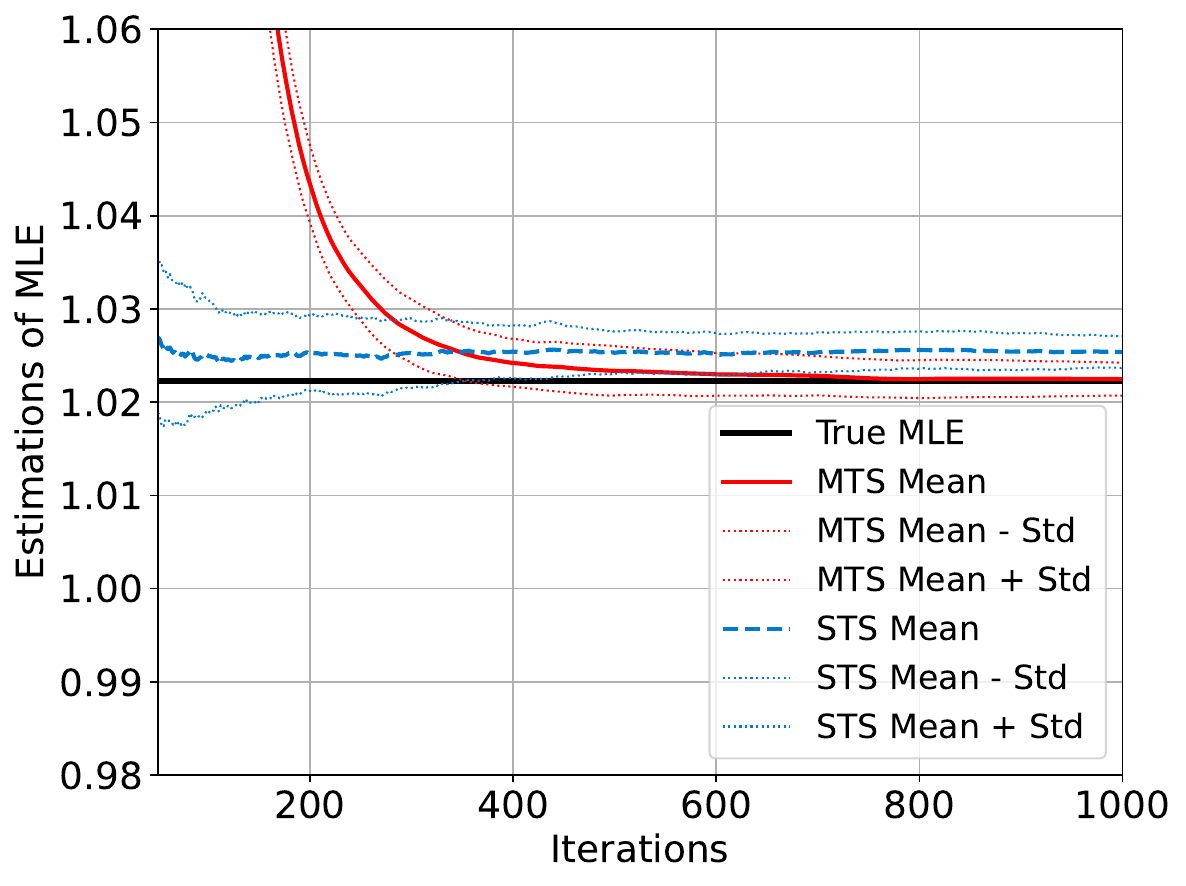}
    \label{ex1_2_2}
  }
\label{ex1_2}
\end{figure}

 The true value $\theta$ is set to be 1. The faster and slower step-size is chosen as $\frac{10}{k^{0.55}}$ and $\frac{0.5}{k}$, respectively, which satisfies the step-size condition of the MTS algorithm. We set $T = 100$ observations, the feasible region $\Theta = [0.5,2]$, and the initial value $\theta_0 = 0.8$. The samples of $X_t = (X_{1,t}, X_{2,t})$ are simulated to estimate the likelihood function and its gradient at each iteration. We compare our MTS algorithm with the STS method. In previous works, a large number of simulated samples per iteration (e.g., $10^5$) is required to ensure a negligible ratio bias from the log-likelihood gradient estimator. By employing our method, computational costs are reduced while improving estimation accuracy. Figure \ref{ex1_2}\subref{ex1_2_1} exhibits the convergence results of MTS and STS  with $N=10^4$ simulated samples based on 100 independent experiments. Compared to the true MLE, MTS achieves lower bias and standard error than STS. The convergence curve is also more stable due to the elimination of the denominator estimator. The average CPU time per experiment for MTS and STS is 0.7s and 0.72s, respectively,  indicating comparable computational costs. Figure \ref{ex1_2}(b) depicts the convergence result with $10^5$ simulated samples based on 100 independent experiments. Even with a large number of simulated samples, MTS outperforms STS. Table \ref{tab:average_bias} records the absolute bias for the two estimators under their respective optimal allocation policies \shortcite{li2024eliminating}, based on 100 independent experiments. $N$ is the batchsize, $K$ is the iteration size, and $\Gamma$ is the total budget. Across all budget levels, MTS demonstrates significantly higher estimation accuracy than STS.
%\vspace{-0.2cm}
%\vspace{-0.5cm}
%According to Theorem \ref{thm10}, the optimal allocation in MTS algorithm for iteration times $K$ and sample size $N$ is $K=O(\Gamma^{\frac{2}{3}})$ and $N=O(\Gamma^{\frac{1}{3}})$, respectively. While for STS algorithm, the optimal allocation is $K=O(\Gamma^{\frac{1}{3}})$ and $N=O(\Gamma^{\frac{2}{3}})$.

\begin{table}[h]
\centering
\caption{The absolute bias of the two estimators and true MLE, based on 100 independent experiments}
\label{tab:average_bias}
\footnotesize
\begin{tabular}{ccccc}
\toprule
\multirow{2}{*}{$\Gamma$} & \multirow{2}{*}{N (K for STS)} & \multirow{2}{*}{K (N for STS)} & \multicolumn{2}{c}{\textbf{Absolute Bias $\pm$ std}}  \\
\cmidrule{4-5}
&&& MTS & STS  \\
\midrule
$10^4$  & 86 & 116   & $1.9\times 10^{-2} \pm 2.2\times 10^{-1}$ & $1.5\times 10^{-1} \pm 3.9\times 10^{-1}$ \\

$3\times 10^4$  & 124 & 241   & $1\times 10^{-2} \pm 8\times 10^{-2}$ & $1\times 10^{-1} \pm 3.8\times 10^{-1}$ \\

$10^5$  & 186 & 539   & $2.3\times 10^{-3} \pm 8\times 10^{-2}$ & $6.4\times 10^{-2} \pm 3.6\times 10^{-1}$ \\

$3\times 10^5$ & 268 & 1120   & $1.5\times 10^{-3} \pm 3.9\times 10^{-2}$  & $2.9\times 10^{-2} \pm 2.8\times 10^{-1}$ \\

$10^6$ & 400 & 2500   & $4.8\times10^{-4} \pm2.2 \times10^{-2}$ & $7.3\times10^{-3} \pm 2.8\times10^{-1}$  \\

$3\times 10^6$ &577 & 5200   & $3\times10^{-4} \pm 1.3 \times10^{-2}$ & $3.4\times10^{-3} \pm 2.6\times10^{-1}$  \\
$10^7 $ & 862 & 11604  & $2\times10^{-4} \pm 8.1 \times10^{-3}$ & $2.1\times10^{-3} \pm 2.9\times10^{-1}$  \\
$3 \times 10^7 $ & 1243 & 24137  & $1.6\times10^{-4} \pm 4.7 \times10^{-3}$ & $1.9\times10^{-3} \pm 1.8\times10^{-1}$  \\
$10^8 $ & 1857 & 53861  & $5.9\times10^{-5} \pm 2.1 \times10^{-3}$ & $1\times10^{-3} \pm 1.2\times10^{-1}$  \\
\bottomrule
\end{tabular}
\end{table}
%\vspace{-0.3cm}

\subsection{PDE Case}\label{sec5.2}
We apply Algorithm \ref{algor:2} to test the nested MTS framework in the PDE setting. Let the prior distribution of the parameter $\theta$ be the standard normal $N(0,1)$. The stochastic model is  $Y_t = X_t + \theta $ with latent variable $X_t\sim N(0,1)$. Given the observation $y=\{Y_t\}_{t=1}^T$, the goal is to compute the posterior distribution for $\theta$. It is straightforward to derive that the analytical posterior is $p(\theta|y) \sim N(\frac{n}{1+n}\bar{y},\frac{1}{1+n}).$

Let the posterior parameter $\lambda$ be $(\mu,\sigma^2)$. We want to use normal distribution $q_{\lambda}(\theta)$ to approximate the posterior of $\theta$,  i.e., $q_{\lambda}(\theta) \sim N(\mu,\sigma^2)$. Applying the re-parameterization technique,  we can sample $u$ from normal distribution $N(0,1)$ and set $\theta(u;\lambda) = \mu + \sigma u \sim N(\mu, \sigma^2)$. Here is just an illustrative example of normal distribution, re-parameterization technique can be applied to other more general distributions \shortcite{figurnov2018implicit,ruiz2016generalized}.

\begin{figure}[h]  
 \centering
 \caption{Trajectories of MTS and STS with sample size $10^4$ based on 100 independent experiments}
 \label{ex2_1}

 \subfigure[Estimations of posterior mean]{
 \centering
 \includegraphics[width=7.5cm]{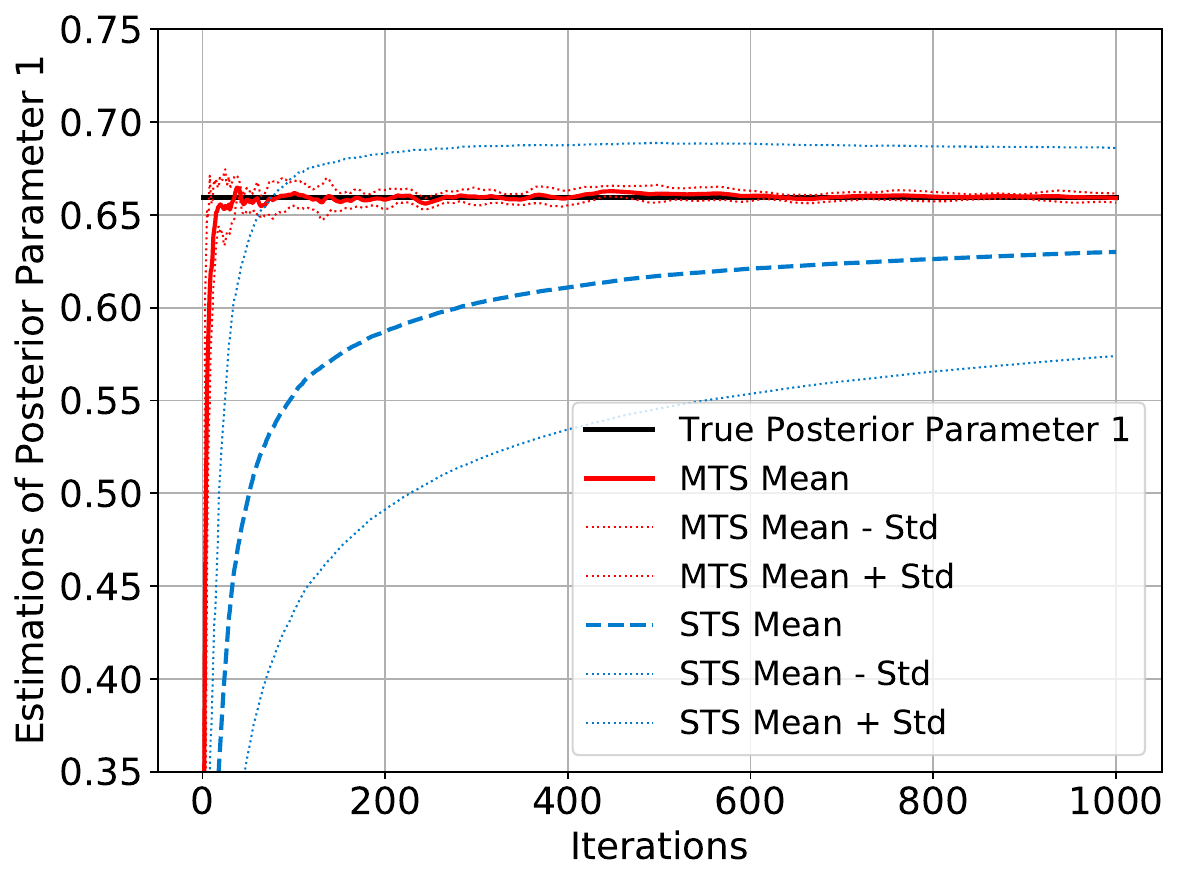}
 \label{ex2_1_1}
 }
 %\hspace{0.5cm}
 \subfigure[Estimations of posterior variance]{
 \centering
 \includegraphics[width=7.5cm]{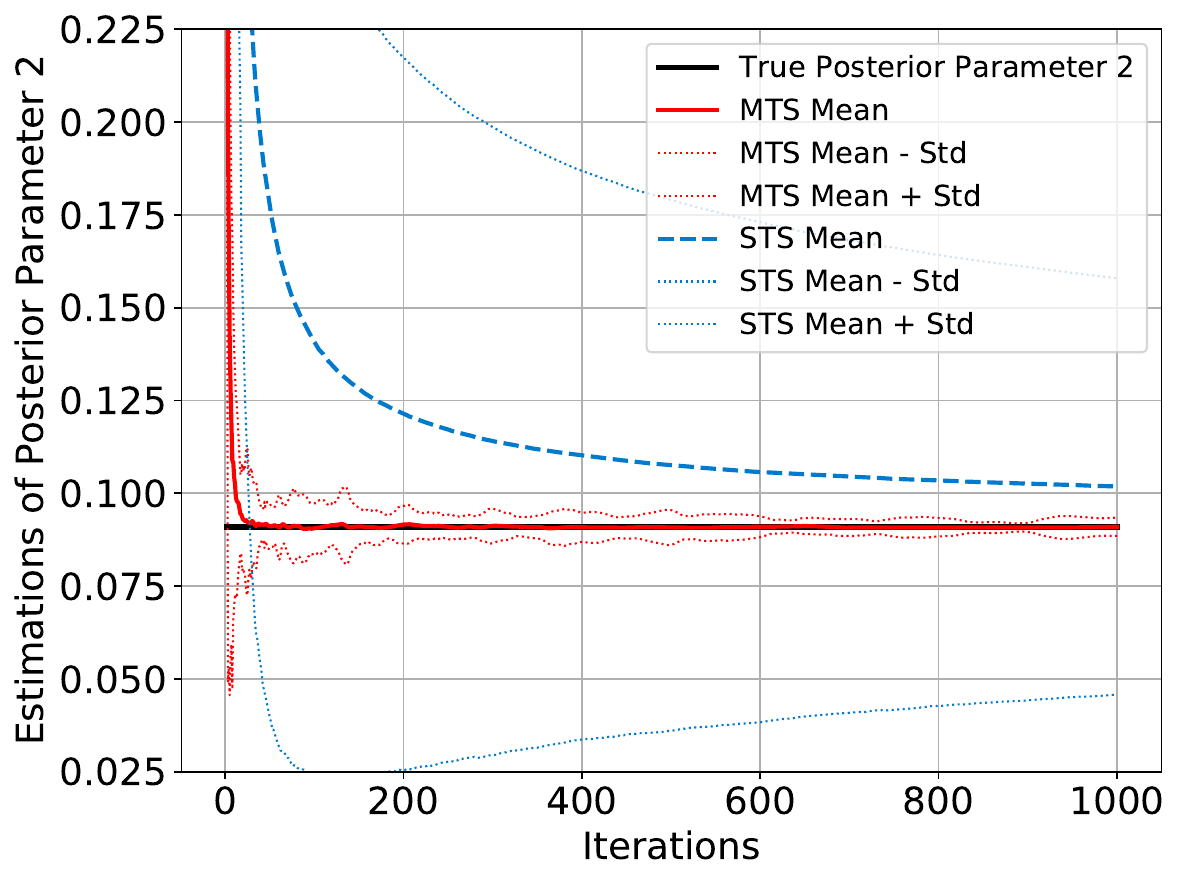}
 \label{ex2_1_2}
 }
\end{figure}

In the PDE case, we can incorporate the data into prior over and over again. Suppose there are only 10 independent observations for one batch. Set feasible region $\Lambda = [-1,10]\times[0.01,2]$ and initial value $\lambda_0= (0,1)$. First, we set $M=10$ outer layer samples $u_m$ and compare the MTS algorithm with the analytical posterior and STS method. The faster and slower step-size is chosen as $\frac{10}{k^{0.55}}$ and $\frac{1}{k}$, respectively. Figure \ref{ex2_1} displays the trajectories of MTS and STS with sample size $10^4$ based on 100 independent experiments. Specifically, Figure \ref{ex2_1}\subref{ex2_1_1} exhibits the convergence for the posterior mean $\mu$ and Figure \ref{ex2_1}\subref{ex2_1_2} exhibits the convergence for the posterior variance $\sigma^2$. MTS achieves lower bias and standard error than STS when compared to the true posterior parameters.

Table \ref{ex2_3} records the absolute error for both estimators under their respective optimal allocation policies, based on 100 independent experiments. Across all budget levels, MTS consistently outperforms STS in estimation accuracy. 
%\vspace{-0.3cm}
\begin{table}[h]
\centering
\caption{The absolute bias of the two estimators, based on 100 independent experiments}
\label{ex2_3}
\footnotesize
\begin{tabular}{cccccccc}
\toprule
\multirow{2}{*}{$\Gamma$} &\multirow{2}{*}{M}  & \multirow{2}{*}{N (K for STS)}  &\multirow{2}{*}{K (N for STS)} & \multicolumn{2}{c}{\textbf{Posterior Mean}} & \multicolumn{2}{c}{\textbf{Posterior Variance}} \\
\cmidrule{5-8}
& & & & MTS & STS & MTS & STS \\
\midrule
$10^5$ & 4 & 214 & 106    & $2.3\times 10^{-3}$ & $8.3\times 10^{-2}$ & $5.5\times 10^{-3}$ & $5.9\times 10^{-2}$ \\

$3\times10^5$ & 5 & 281 & 183   & $1.1\times 10^{-3}$  & $4.5\times 10^{-2} $ & $2.8\times 10^{-3}  $ & $2\times 10^{-2}$\\

$10^6$ & 7 & 380 & 334   & $5.2\times 10^{-4}$  & $1.7\times 10^{-2} $ & $6.8\times 10^{-4}  $ & $6.2\times 10^{-3}$\\

$3\times10^6$ & 10 & 500 & 578   & $3.0\times10^{-4} $ & $1.2\times10^{-2} $ & $1.0\times10^{-4} $ & $2.3\times10^{-3}$ \\

$10^7$ & 14 & 675 & 1055   & $1.9\times 10^{-4}$  & $5\times 10^{-3} $ & $9.4\times 10^{-5}  $ & $6\times 10^{-4}$\\

$3\times10^7$ & 18 & 889 & 1826   & $5\times10^{-5} $ & $4.1\times10^{-3} $ & $4.9\times10^{-5} $ & $3.7\times10^{-4}$ \\

$10^8$ & 25 & 1200 & 3334   & $4.2\times 10^{-5}$  & $2.3\times 10^{-3} $ & $1.5\times 10^{-5}  $ & $2.7\times 10^{-4}$\\

$3\times10^8$ &32 & 1580 & 5774   & $1.3\times10^{-5} $ & $1.4\times10^{-3} $ & $5.2\times10^{-6} $ & $8.1\times10^{-5}$ \\

$10^9$ &44 & 2134 & 10561   & $1\times10^{-5} $ & $8.1\times10^{-4} $ & $6.6\times10^{-6} $ & $4.2\times10^{-5}$ \\
\bottomrule
\end{tabular}
\end{table}
%\vspace{-0.3cm}

% Figure \ref{ex2_2} illustrates the log-log plot of the MAE of the estimators versus  $\Gamma$ in 1000 different experimental settings. For each setting, MTS and STS are run once. Every experiment is done under the aforementioned optimal budget allocation policy in corresponding algorithms. MTS consistently outperforms STS in all scenarios.  %The convergence rate $O(\Gamma^{-\frac{1}{8}})$ in the theoretical result is not as better as the experimental result, where the rate is still around $O(\Gamma^{-\frac{1}{3}})$. 
% \begin{figure}[h]
%   \centering
%   \caption{Log-log plot of the MAE of the estimators versus the total budget $\Gamma$ of MTS and STS algorithm based on 1000 different experimental settings}
%   \label{ex2_2}

%   \subfigure[Convergence rate of posterior mean]{
%     \centering
%     \includegraphics[width=6cm]{ex2_2_mu_rate.pdf}
%     \label{ex2_2_1}
%   }
%   %\hspace{0.5cm}
%   \subfigure[Convergence rate of posterior std]{
%     \centering
%     \includegraphics[width=6cm]{ex2_2_sigma_rate.pdf}
%     \label{ex2_2_2}
%   }

% \end{figure}
\subsection{MLE for the HMM}\label{5.3}
We illustrate our approach through the following example. The initial value of the hidden state is set as $S_0 = 0$. The transition kernel is defined as $S_t = S_{t-1} + \theta + V_t$, and the observation kernel is given by $Y_t = S_t + W_t$, where $S_t$ denotes the hidden state, while $W_t$ and $V_t$ represent the observation error and transition error, respectively. Here, $W_t \sim N(0,1)$ and $V_t \sim N(0,1)$, and both $W_t$ and $V_t$ are mutually independent and identically distributed (i.i.d.). Our goal is to estimate the parameter $\theta$ using the observations $\{Y_t\}_{t=1}^T$.

In this example, we set $T = 100$ and use $J = 10^3$ or $10^4$ particles. The step sizes are chosen as $\frac{100}{K^{0.8}}$ and $\frac{0.1}{K}$, respectively. We conduct 20 independent experiments, replacing the GLR estimators with direct computation of particle weights using the observation kernel. By comparing the results presented in Table \ref{tab1}, we conclude that our method significantly reduces bias in the HMM.
\begin{table}[h!]
\centering
\caption{The mean absolute error of the two estimators based on 20 independent experiments}
\label{tab1}
\footnotesize
\begin{tabular}{cccc}
\toprule
 \multirow{2}{*}{Particle Numbers} & \multicolumn{2}{c}{\textbf{Mean Absolute Error}} \\
\cmidrule{2-3}
& MTS & STS \\
\midrule
 100  & 0.0307 & 0.0427 \\
 1000 & 0.0104 & 0.0145 \\
\bottomrule
\end{tabular}
\end{table}
\section{Conclusion}\label{sec6}
This paper presents a comprehensive study addressing the challenges of parameter estimation where the likelihood function is estimated by simulations. Our GSPE approach, grounded in the MTS algorithm, handles the ratio bias problem,  enhances the accuracy of parameter estimation, and saves computational costs. In the realm of PDE, we have explored a nested simulation optimization structure, which is both theoretically sound and empirically effective. 
% Reducing font size (to 9pt) for References & Author Biagraphies
\section*{ACKNOWLEDGMENTS}
This work was supported in part by the National Natural Science Foundation of China (NSFC) under
 Grants 72325007, 72250065, and 72022001. %as well as the Special Funds for Guiding Local Scientific and
 %Technological Development by the Central Government under Grant 2023EGA035.

\footnotesize

% Please don't exchange the bibliographystyle style
\bibliographystyle{wsc}

% AUTHOR: Include your bib file here
\bibliography{wsc25paper}

\section*{AUTHOR BIOGRAPHIES}

\noindent {\bf \MakeUppercase{Zehao Li}} is a PhD candidate in the Department of Management Science and Information Systems in Guanghua
 School of Management at Peking University, Beijing, China. He received the BS degree in  School of Mathematical Sciences, Fudan University in 2023. His research interests include simulation optimization and machine learning. His email address is \email{zehaoli@stu.pku.edu.cn}.\\

\noindent {\bf \MakeUppercase{Yijie Peng}}  is an Associate Professor in Guanghua School of Management at Peking University, Beijing, China. His research
 interests include stochastic modeling and analysis, simulation optimization, machine learning, data analytics, and healthcare.  He is a member of
 INFORMS and IEEE and serves as an Associate Editor of the Asia-Pacific Journal of Operational Research and the Conference
 Editorial Board of the IEEE Control Systems Society. 
 His email address is  \email{pengyijie@pku.edu.cn}.\\

\newpage

\end{document}